\documentclass[twoside,11pt]{article}

\usepackage[accepted,specialissue]{melba}

\usepackage{amsmath,amsfonts}
\usepackage[nolist,nohyperlinks]{acronym}
\usepackage[capitalize,noabbrev]{cleveref}
\usepackage{caption}
\usepackage{subcaption}
\usepackage{placeins}
\usepackage{booktabs}
\usepackage{graphicx}

\acrodef{AI}{artificial intelligence}
\acrodef{ML}{machine learning}
\acrodef{XAI}{explainable artificial intelligence}
\acrodef{LR}{logistic regression}
\acrodef{CLAHE}{Contrast Limited Adapted Histogram Equalization}
\acrodef{TCAV}{Testing with Concept Activation Vectors}
\acrodef{CAV}{concept activation vector}
\acrodef{CBM}{Concept Bottleneck Model}
\acrodef{coop-CBM}{cooperative-Concept Bottleneck Model}
\acrodef{CRP}{Concept Relevance Propagation}
\acrodef{DR}{diabetic retinopathy}
\acrodef{MCC}{Matthews correlation coefficient}
\acrodef{MA}{microaneurysms}
\acrodef{HE}{hemorrhages}
\acrodef{EX}{hard exudates}
\acrodef{SE}{soft exudates}
\acrodef{IRMA}{intraretinal microvascular abnormalities}
\acrodef{NV}{neovascularization}
\acrodef{FGADR}{fine-grained annotations diabetic retinopathy}
\acrodef{APTOS}{Asia Pacific Tele-Ophthalmology Society}
\acrodef{IDRiD}{Indian Diabetic Retinopathy Image Dataset}

\melbaid{2024:021}  
\doi{https://doi.org/10.59275/j.melba.2024-e7fd}
\melbaauthors{Stor{\aa}s and Sundgaard}  
\volume{2}
\firstpageno{2053}  
\melbayear{2024}  
\datesubmitted{03/2024}  
\datepublished{10/2024}  

\melbaspecialissue{Interpretability of Machine Intelligence in Medical Image Computing \\ (iMIMIC) 2023}
\melbaspecialissueeditors{Mauricio Reyes, Jaime Cardoso, Jayashree Kalpathy-Cramer, Nguyen Le Minh, Pedro Abreu, José Amorim, Wilson Silva, Mara Graziani, Amith Kamath}

\ShortHeadings{Looking into Concept Explanation Methods for DR Classification}{Stor{\aa}s and Sundgaard}

\title{Looking into Concept Explanation Methods for \\ Diabetic Retinopathy Classification}

\author{\firstname Andrea M. \surname Stor{\aa}s\orcid{0000-0002-1038-7080} \email btis@novonordisk.com \\  
	\addr Department of Holistic Systems, Simula Metropolitan Center for Digital Engineering, Oslo, Norway and \\ Novo Nordisk A/S, Søborg, Denmark
	\AND
	\firstname Josefine V. \surname Sundgaard\orcid{0000-0003-2872-4660} \email jfvs@novonordisk.com \\
	\addr Novo Nordisk A/S, Søborg, Denmark and \\ Department of Applied Mathematics and Computer Science, Technical University of Denmark, Kongens Lyngby, Denmark
}

\begin{document}

\maketitle


\begin{abstract}
	Diabetic retinopathy is a common complication of diabetes, and monitoring the progression of retinal abnormalities using fundus imaging is crucial. Because the images must be interpreted by a medical expert, it is infeasible to screen all individuals with diabetes for diabetic retinopathy. Deep learning has shown impressive results for automatic analysis and grading of fundus images. One drawback is, however, the lack of interpretability, which hampers the implementation of such systems in the clinic. Explainable artificial intelligence methods can be applied to explain the deep neural networks. Explanations based on concepts have shown to be intuitive for humans to understand, but have not yet been explored in detail for diabetic retinopathy grading. This work investigates and compares two concept-based explanation techniques for explaining deep neural networks developed for automatic diagnosis of diabetic retinopathy: Quantitative Testing with Concept Activation Vectors and Concept Bottleneck Models. We found that both methods have strengths and weaknesses, and choice of method should take the available data and the end user's preferences into account.
	Our code is available at~\url{https://github.com/AndreaStoraas/ConceptExplanations\_DR\_grading}.
\end{abstract}

\begin{keywords}
	Explainable Artificial Intelligence, Concept-Based Explanations, Diabetic Retinopathy, Fundus Images
\end{keywords}

\section{Introduction}
\begin{figure}[t!]
     \centering
     \begin{subfigure}[b]{0.19\textwidth}
         \centering
         \includegraphics[width=1\textwidth]{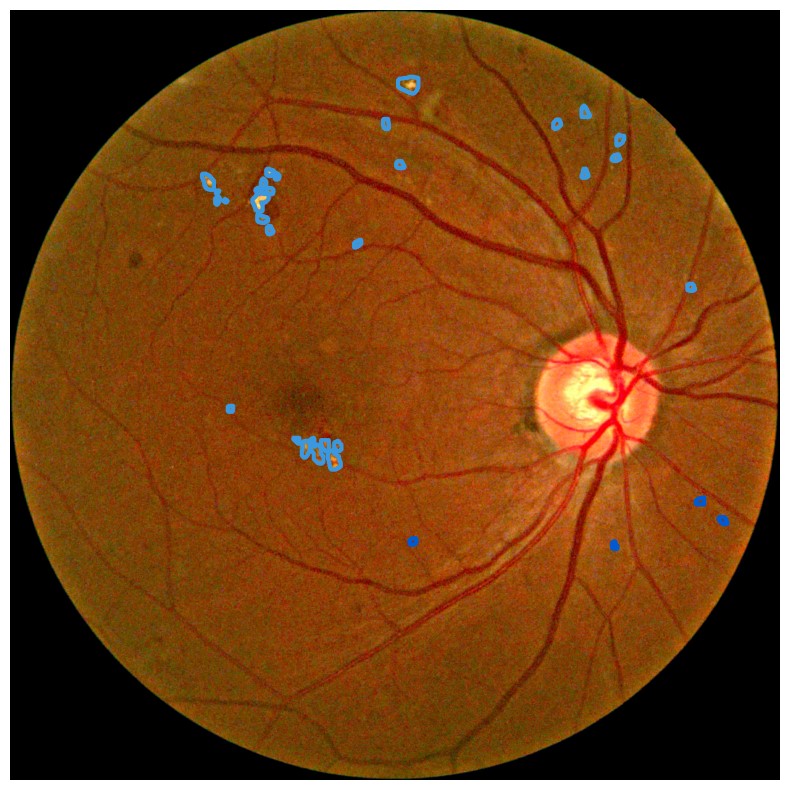}
         \caption{\label{fig:DRLevel0} \ac{DR} level 0}
     \end{subfigure}
     \hfill
     \begin{subfigure}[b]{0.19\textwidth}
         \centering
         \includegraphics[width=1\textwidth]{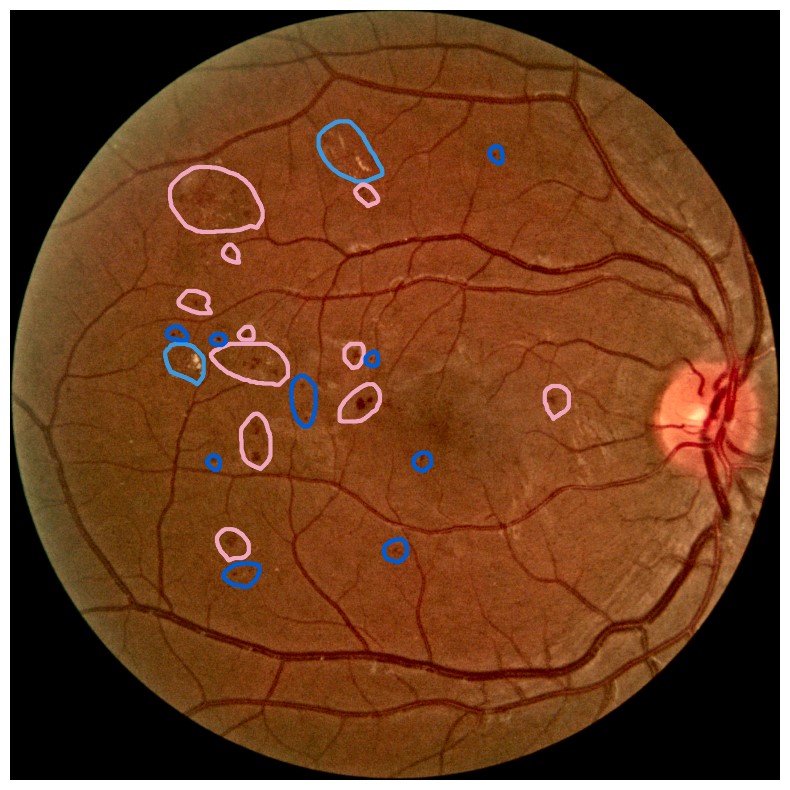}
         \caption{\label{fig:DRLevel1} \ac{DR} level 1}
    \end{subfigure}
     \hfill
     \begin{subfigure}[b]{0.19\textwidth}
         \centering
         \includegraphics[width=1\textwidth]{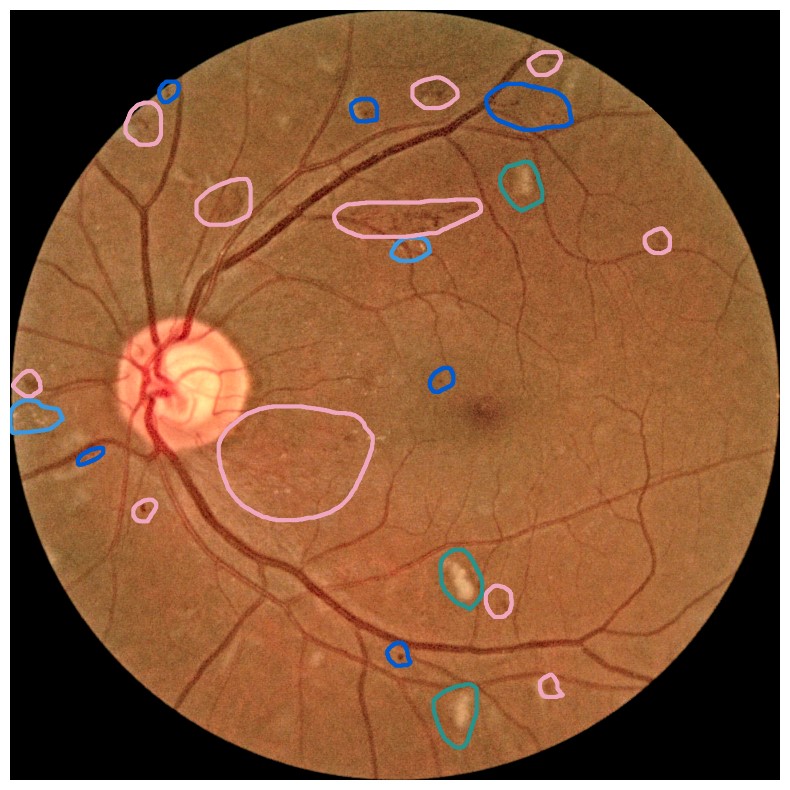}
         \caption{\label{fig:DRLevel2} \ac{DR} level 2}
    \end{subfigure}
     \hfill
     \begin{subfigure}[b]{0.19\textwidth}
         \centering
         \includegraphics[width=1\textwidth]{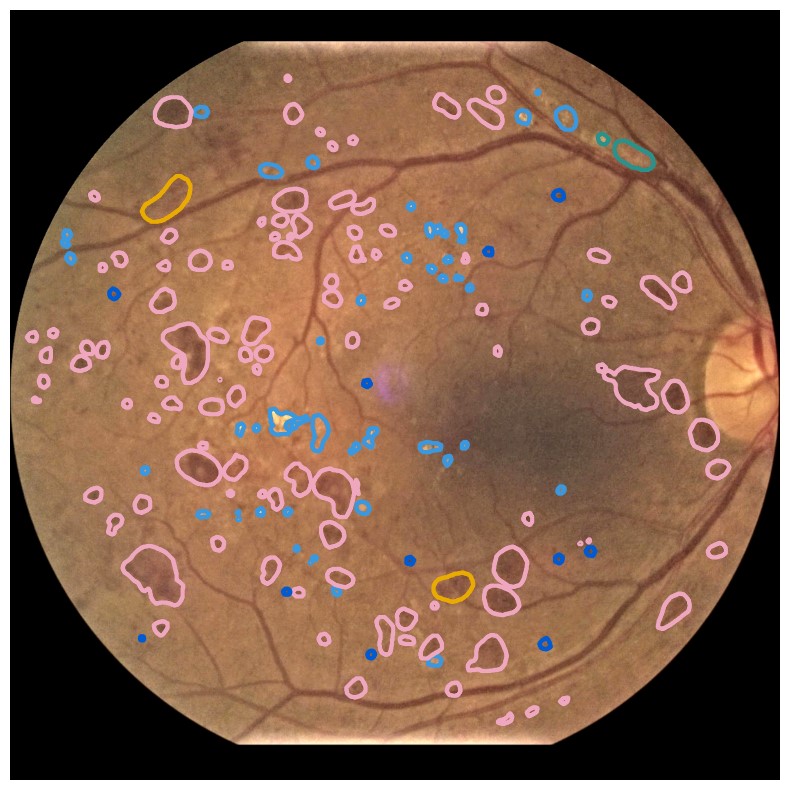}
         \caption{\label{fig:DRLevel3} \ac{DR} level 3}
    \end{subfigure}
     \hfill
     \begin{subfigure}[b]{0.19\textwidth}
         \centering
         \includegraphics[width=1\textwidth]{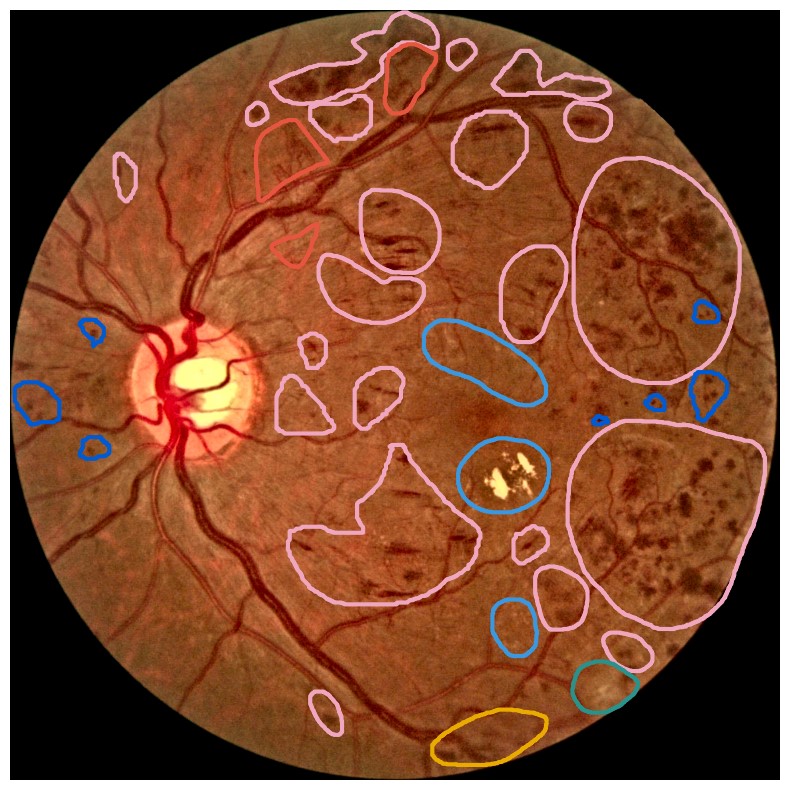}
         \caption{\label{fig:DRLevel4} \ac{DR} level 4}
    \end{subfigure}
    \caption{Example fundus images representing increasing \ac{DR} severity with segmentation masks of retinal lesions. Level 4 is the most severe type of \ac{DR} and is associated with a high risk of blindness. Images from the FGADR dataset~\citep{Zhou2021FGADR}. Dark blue = microaneurysms, pink = hemorrhages, light blue = hard exudates, green = soft exudates, yellow = intra-retinal microvascular abnormalities, and red = neovascularization. Best viewed with zoom.}
    \label{fig:DRLevels}
\end{figure}

Diabetes is a disease with increasing prevalence, and \ac{DR} is one of the most common complications~\citep{WHO_diabetes}. \Ac{DR} is characterized by retinal abnormalities, which damage the eye and can lead to blindness. Its severity depends on the type and amount of retinal abnormalities: \Ac{HE}, \ac{MA}, \ac{EX}, \ac{SE}, \ac{IRMA}, and \ac{NV}. \Ac{DR} is graded from $0$ to $4$ (no \ac{DR}, mild, moderate, and severe nonproliferative \ac{DR}, and proliferative \ac{DR}), as described by~\cite{Wilkinson2003DRgrading}.
\ac{DR} grading of fundus images is traditionally a manual process requiring medical expertise. \Cref{fig:DRLevels} provides examples of fundus images of eyes with increasing severity of \ac{DR} including the ground truth segmentations of the six retinal abnormalities. These images underline the challenge of identifying relevant medical abnormalities without specialist training. 

Deep neural networks have shown impressive results for objectively predicting levels of \ac{DR} from fundus images~\citep{Lakshminarayanan2021DRReview}.
However, these models are complex and difficult to interpret, which is regarded as an obstacle for clinical implementation of such systems~\citep{vellido2020XAIimportance}. If ophthalmologists do not understand why the model made the specific predictions, they might refuse to use it. Moreover, if algorithms make decisions that can affect the life of the patient to a large extent, such as in medical diagnoses, the patient has the right to get an explanation about why the decision was made.
In other words, being able to explain the decision process of deep neural networks for medical applications is crucial. 

\Ac{XAI} aims to explain machine learning models and their predictions. Previous work on \ac{XAI} for fundus image analysis have mainly focused on heatmap methods, which highlight the pixels in the image regarded as important during inference~\citep{vanDerVelden2022XAI_surveyFundus}. As an example, \cite{Ayhan2022DRHeatmapEval} performed a quantitative analysis of heatmaps produced by a wide selection of \ac{XAI} methods to explain deep learning-based \ac{DR} detection in fundus images, showing that the quality of heatmaps vary greatly. Despite their popularity, heatmap methods have some limitations~\citep{Arun2021SaliencyMapsNotWorking}. Heatmaps do not explain why an area in the image was highlighted~\citep{Kovalerchuk2021HeatmapsNotExplainingWhy}, or quantify how important the highlighted area is. Moreover, one heatmap is generated for each observation, making it challenging to get a global overview of how the model interprets images. 

Concept explanation methods are attractive for medical applications because they measure how much the deep neural networks are influenced by high-level concepts representing clinical findings~\citep{Salahuddin2022interpretabilitySurvey}. A concept can be described as a theme or topic, e.g., 'stripes' and 'dots' for natural images, or diagnostic findings such as 'hemorrhages' and 'microaneurysms' for fundus images. The six different diagnostic concepts used for this work is shown with the ground truth segmentation masks in~\cref{fig:DRLevels}. \Ac{TCAV}~\citep{been2018TCAV} and \acp{CBM}~\citep{Koh2020BottleneckModels} are two concept-based methods that have several advantages above heatmap methods. Both allow the user to define the concepts, which ensures relevant and meaningful concepts that are interpretable for the end-users. How the concepts are used varies between the two explanation methods and is outlined in~\cref{Sec:Method_TCAV,Sec:Method_CBM}. Moreover, the relative importance of the concepts can be quantitatively measured. For \ac{TCAV}, concept scores can be generated for a group of images, e.g. images belonging to the same class, allowing the user to investigate whether the model has learnt aspects coherent to domain knowledge and diagnostic guidelines. \acp{CBM}, on the other hand, allow the user to directly modify the model's intermediate concept predictions at test time and observe how this affects the final model prediction. This way of manipulating the model after training is attractive in the medical field, e.g. if the clinician wants to increase the emphasis of a concept in the image that the model missed. While \ac{TCAV} explains models \textit{post-hoc}, i.e., predicting the concepts after the classification model has been trained, \acp{CBM} provide \textit{ad-hoc} explanations, where the prediction of the concepts are trained together with the classification model.  
Even though concept explanations can be more intuitive than heatmaps for medical doctors, neither \ac{TCAV} or \acp{CBM} have been extensively studied in the field of \ac{DR} grading. In this work, we thus investigate and compare \ac{TCAV} and \acp{CBM} for explaining deep neural networks trained to grade \ac{DR} in fundus images. 

\section{Data and Method}
Seven open access datasets were used in the current study: APTOS~\citep{APTOS2019-dataset}, DR Detection~\citep{DRDetection_dataset,Cuadros2009eyePACS}, Messidor-2~\citep{Decenciere2014Messidor,Abramoff2013Messidor2}, FGADR~\citep{Zhou2021FGADR}, DDR~\citep{Li2019DDR_dataset}, DIARETDB1~\citep{Kauppi2007diaretdb1} and IDRiD~\citep{IDRiD2018_datasetCitation}. An overview of the data is provided in~\cref{tab:datasets}.
All datasets including image-level annotations of \ac{DR} grade were used to train the deep neural networks for \ac{DR} grading. Several data sources are combined to ensure generalizable models, as the fundus images were captured at different locations, by different healthcare personnel, and using different devices, making the training data diverse.
For concept generation, fundus images segmented with medical findings relevant for diagnosis of \ac{DR} were used. FGADR was applied for both \ac{DR} grading and concept generation. The distribution of \ac{DR} levels are highly skewed, with the majority of the images representing eyes with no signs of \ac{DR}, and annotations of \ac{IRMA} and \ac{NV} are only available in FGADR. The four datasets used for developing the \ac{DR} grading models were split into $80 \%$ for training, $10 \%$ for validation, and $10 \%$ testing. Images from the same patient were placed in the same split of the dataset. Additionally, the black background was removed from all images. Further on, \ac{CLAHE} was applied to enhance the image quality by making the blood vessels and retinal findings more visible~\citep{zuiderveld1994clahe}. For the training set, several image augmentation techniques, such as random flipping, blurring, and change of perspective, were also applied. The source code for all the experiments, including concept explanations, is publicly available online\footnote{\url{https://github.com/AndreaStoraas/ConceptExplanations\_DR\_grading}}.

\begin{table*}[t!]
    \caption{Description of the applied datasets. NA: Not available.} 
    \centering
    {\begin{tabular}{l c c c c c c }
        \toprule
         Datasets with \ac{DR} grading & Total & Level 0 & Level 1 & Level 2 & Level 3 & Level 4 \\
        \midrule 
        APTOS & $3662$ & $1805$ & $370$ & $999$ & $193$ & $295$ \\
        DR Detection
         & $35126$ & $25810$ & $2443$ & $5292$ & $873$ & $708$ \\
         Messidor-2 & $1744$ & $1017$ & $270$ & $347$ & $75$ & $35$ \\
        FGADR & $1842$ & $101$ & $212$ & $595$ & $647$ & $287$ \\
        \midrule
        Datasets with segmentations & Total & MA & HE & SE & EX & IRMA/NV \\
        \midrule
        FGADR & $1842$ & $1424$ & $1456$ & $627$ & $1279$ & $159$/$49$ \\
        DDR & $757$ & $570$ & $601$ & $239$ & $486$ & NA \\
        DIARETDB1 & $89$ & $80$ & $54$ & $36$ & $48$ & NA \\
        IDRiD & $81$ & $81$ & $80$ & $40$ & $81$ & NA \\
        \bottomrule
    \end{tabular}}
    \label{tab:datasets}
\end{table*}

\subsection{Testing with Concept Activation Vectors}\label{Sec:Method_TCAV}
Two model architectures, Inception V3~\citep{Szegedy2016InceptionV3} and Densenet-121~\citep{Huang2017Densenet121}, were applied due to good performance on analyzing fundus images in previous work~\citep{Kora2022TL_medicine,Zoi2022multimodalTL,Zhou2021FGADR}.
Both models were pretrained on ImageNet~\citep{Deng2009Imagenet}, as also used in the previous works, and modified to predict five classes in the final prediction layer. The models were fine-tuned for $100$ epochs on the combined training and validation sets for \ac{DR} grading.
The best performing model on the validation set was used for further evaluation. Due to class imbalance, a weighted random sampler was used during training. Moreover, an Adam optimizer with default hyperparameters and cross-entropy loss were applied~\citep{Paszke2019Pytorch}. 

\ac{TCAV} measures the relative concept importance for a classification result by checking how sensitive the model is to changes in the input image toward the direction of the concept, defined by the concept's corresponding \ac{CAV}~\citep{been2018TCAV}. For estimating the \ac{CAV}, the user provides a set of positive example images containing the concept of interest and a set of negative example images, where the concept is absent. Next, features are extracted from the positive and negative example sets, respectively, from a specific layer in the model chosen by the user. A linear classifier is trained to separate the features from the two image sets apart, and the resulting \ac{CAV} lies orthogonal to the classification boundary of the linear model. The process is repeated for each concept. For more details about the mathematics behind the \ac{XAI} method, the interested reader is referred to the original \ac{TCAV} paper~\citep{been2018TCAV}. In our work, the concepts were defined as described in the original publication~\citep{been2018TCAV}. First, the full images were used to represent the concepts~\citep{been2018TCAV}, as opposed to cropping out the image regions where the specific medical finding for a given concept were located. However, abnormalities in fundus images can be small, and are typically not evenly distributed in the image. We therefore tested a second way of preprocessing the concept images, masking out the area around the relevant medical findings based on the segmentation masks. This might enhance the quality of the concepts and was inspired by~\cite{chen2020conceptCrop}. To avoid extreme variations in image sizes, the masked images were restricted to be at least $520\times520$ pixels. For both concept generation approaches, \ac{CLAHE} was applied to enhance the quality of the concept images.

Regardless of the image preprocessing approach, concepts were generated for all six medical findings used for grading \ac{DR} in fundus images (\ac{MA}, \ac{HE}, \ac{EX}, \ac{SE}, \ac{IRMA}, and \ac{NV}) based on the segmentation masks. Positive examples containing the concept and negative examples without the concept were employed for generating the concept activation vectors. The presence of other findings in the images were balanced between the positive and negative examples. The positive and negative example sets contained $45$ images each, which were randomly selected from the four datasets. To test the significance of the detected concepts, $20$ different negative sets were generated for each concept. FGADR was the only dataset annotated with \ac{IRMA} and \ac{NV}, and all positive and negative examples for these two concepts thus arrived from the FGADR dataset.

Next, \ac{TCAV} scores were calculated for images from the test set. The concepts were extracted from Denseblock4, which is the last block before the prediction layer of the Densenet-121 model. To make sure the concepts were consistent and not only caused by randomness, two-sided paired t-tests were performed on the \ac{TCAV} scores for a given \ac{DR} level using the positive example set for a given concept and the $20$ negative example sets.
Only statistically significant concepts with p-values $<0.05$ were considered. 

\subsection{Concept Bottleneck Models}\label{Sec:Method_CBM}

 \begin{figure*}[t]
     \centering \includegraphics[width=\textwidth]{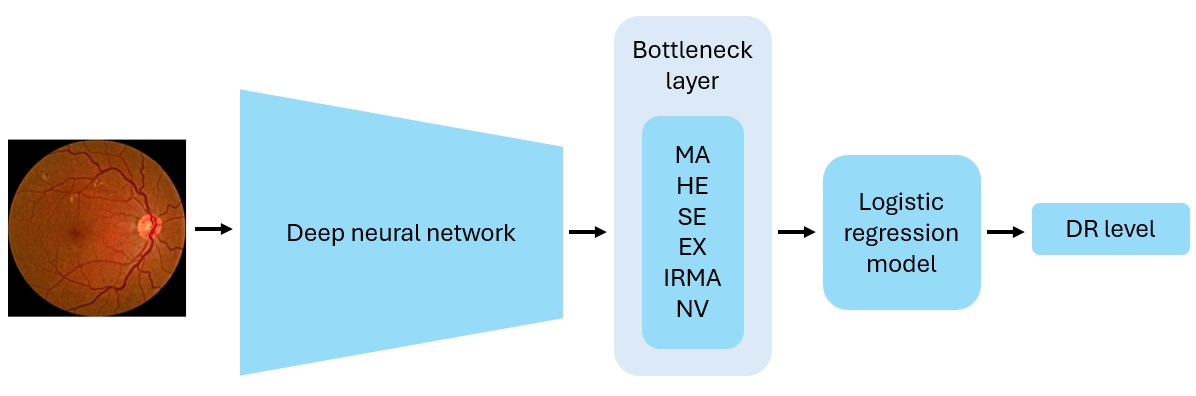}
     \caption{Schematic representation of a sequential bottleneck model predicting DR level from six concepts. The `bottleneck layer' is the concepts predicted by a deep neural network. The predicted concepts are then provided to a logistic regression model for \ac{DR} level classification.}
     \label{fig:Schematic_sequentialBottleneck}
 \end{figure*}
 
\acp{CBM} differ from \ac{TCAV} in that the concepts are learnt together with the target labels during model training. The deep neural network is modified to include a bottleneck layer that predicts the concepts before the final prediction layer~\citep{Koh2020BottleneckModels}. In other words, the final labels are predicted directly from the concepts. By inspecting the probabilities for the predicted concepts for a given input image, it is possible to observe how much each concept influences the model. A drawback of \acp{CBM} is that the dataset must include annotations for both the target labels and the concepts of interest for all images in the dataset. Consequently, getting enough annotated data could be an issue, especially in the medical field, where annotations are typically challenging and expensive to obtain. Moreover, only concepts included during the training phase can be explored. For \ac{TCAV}, this is less of a concern because the concepts are defined after training the model and the example images are not restricted to images from the training dataset.

Because \acp{CBM} learn the concepts during model training, the deep neural network used for \ac{TCAV} could not be used. The original \ac{CBM} paper~\citep{Koh2020BottleneckModels} describes several alternative ways of designing \acp{CBM}. In our experiments, we implemented sequential bottleneck models using a modified version of the Densenet-121 and Inception V3 architectures, as shown in \Cref{fig:Schematic_sequentialBottleneck}. The bottleneck model was initiated with the weights from the corresponding \ac{DR} grading model and fine-tuned to predict the presence of six diagnostic concepts. Next, we trained a \ac{LR} model to classify \ac{DR} levels from the concept predictions provided by the bottleneck model.
The FGADR dataset, being the only dataset annotated with both \ac{DR} levels and all six concepts, was applied for training.
Due to limited amount of training data, another bottleneck model was also trained to only predict the four most prevalent concepts: \ac{MA}, \ac{HE}, \ac{EX}, and \ac{SE}. By excluding the \ac{IRMA} and \ac{NV} concepts, the DDR, IDRiD, and DIARETDB1 datasets can also be used for training. Images of \ac{DR} level 0 without any of the concepts were also included in the training, validation and test sets, encouraging the models to not always predict the most prevalent concepts to be present. Following~\cite{Koh2020BottleneckModels}, binary cross entropy with logits loss was used for training the bottleneck models. Apart from that, the bottleneck and \ac{LR} models were trained with the same hyperparameters as the \ac{DR} grading models.

The main advantage with \acp{CBM} is the possibility of manually correcting the predicted concepts provided to the \ac{LR} model at test time.
Inspired by the original paper~\citep{Koh2020BottleneckModels}, we intervened on the concepts by using the 1st and 99th percentiles for the predicted concept values on the training dataset. These percentiles functioned as surrogates for the true concept values for the absence and presence of a given concepts, respectively. Test time intervention was performed on the entire FGADR test set. Additionally, the intervention was performed on the subset of test images classified with incorrect \ac{DR} levels to make it easier to observe the differences in model performance with and without test time intervention. In both cases, only incorrect concept predictions were corrected using the percentile values. 
The effect of incrementally correcting more concepts was studied, where the concepts were ordered based on the increase in balanced accuracy when adjusting the concepts independently. 

\section{Results}
The models based on the Densenet-121 architecture outperformed the Inception V3-based models for both \ac{TCAV} and \acp{CBM} models. The performance metrics on \ac{DR} grading for the models based on Densenet-121 on the combined and FGADR test sets are presented in~\cref{tab:resultsCombined}. Because concepts were not used when training the models explained by \ac{TCAV}, the performance metrics reported for the `\ac{TCAV} models' can be regarded as baseline results for \ac{DR} level classification. We observe from~\cref{tab:resultsCombined} that the model used for \ac{TCAV} had the highest performance on the combined test set. The \ac{CBM} trained on six concepts generalized poorly from FGADR to the combined test set, but performed best on the FGADR test set. This is not surprising, as the FGADR training set is much smaller and less diverse than the combined training set. For the model trained on four concepts, the performance was not significantly different between the two test sets because of the mixed training data. Regarding the concept detection task, the \ac{CBM} trained on six concepts correctly identified 86.2\% of the diagnostic concepts, compared to 82.5\% for the model based on four concepts. The balanced accuracy also increased from 80.9\% to 85.9\% when increasing the number of concepts. These results were computed on the FGADR test set. The \ac{CBM} trained on six concepts was used in further experiments since this model performed best on the FGADR data for both \ac{DR} grading and concept detection.

\begin{table*}[t!]
    \caption{Performance metrics on both test sets. Highest performance marked in \textbf{bold}. Acc.: Accuracy, MCC: Matthews correlation coefficient, TTI: Test time intervention.}
    \centering
    \begin{tabular}{l c  c  c  c c  c }
        \toprule
         Model & \begin{tabular}{@{}c@{}}No. of \\ concepts\end{tabular} & Acc. & \begin{tabular}{@{}c@{}}Balanced \\ accuracy\end{tabular}  & F1 score & MCC & Precision \\
        \midrule
        Combined test set\\
        \midrule
        TCAV & - & \textbf{81.2\%} & \textbf{62.3\%} & \textbf{0.612} & \textbf{0.615} & \textbf{0.613} \\
        CBM & $4$ & $71.9\%$ & $44.8\%$ & $0.429$ & $0.416$  & $0.454$ \\
        CBM & $6$ & $24.8\%$ & $39.9\%$ & $0.257$ & $0.095$ & $0.318$  \\
        \midrule
        FGADR test set\\
        \midrule
        TCAV & - & \textbf{66.7\%} & $55.2\%$ & $0.529$ & \textbf{0.547} & $0.511$  \\
        CBM & $4$ & $52.9\%$ & $51.7\%$ & $0.470$ & $0.384$  & $0.461$\\ 
        CBM & $6$ & $55.0\%$ & $56.0\%$ & $0.521$ & $0.416$ & $0.525$  \\
        CBM + TTI (full) & $6$ & $54.0\%$ & $59.0\%$ & $0.532$ & 0.412 & $0.532$  \\
        CBM + TTI (incorrect) & $6$ & $64.6\%$ & \textbf{69.4\%} & \textbf{0.634} & $0.545$ & \textbf{0.628} \\
        \bottomrule
    \end{tabular}
    \label{tab:resultsCombined}
\end{table*}

\begin{figure*}[!bt]
     \centering \includegraphics[width=\textwidth]{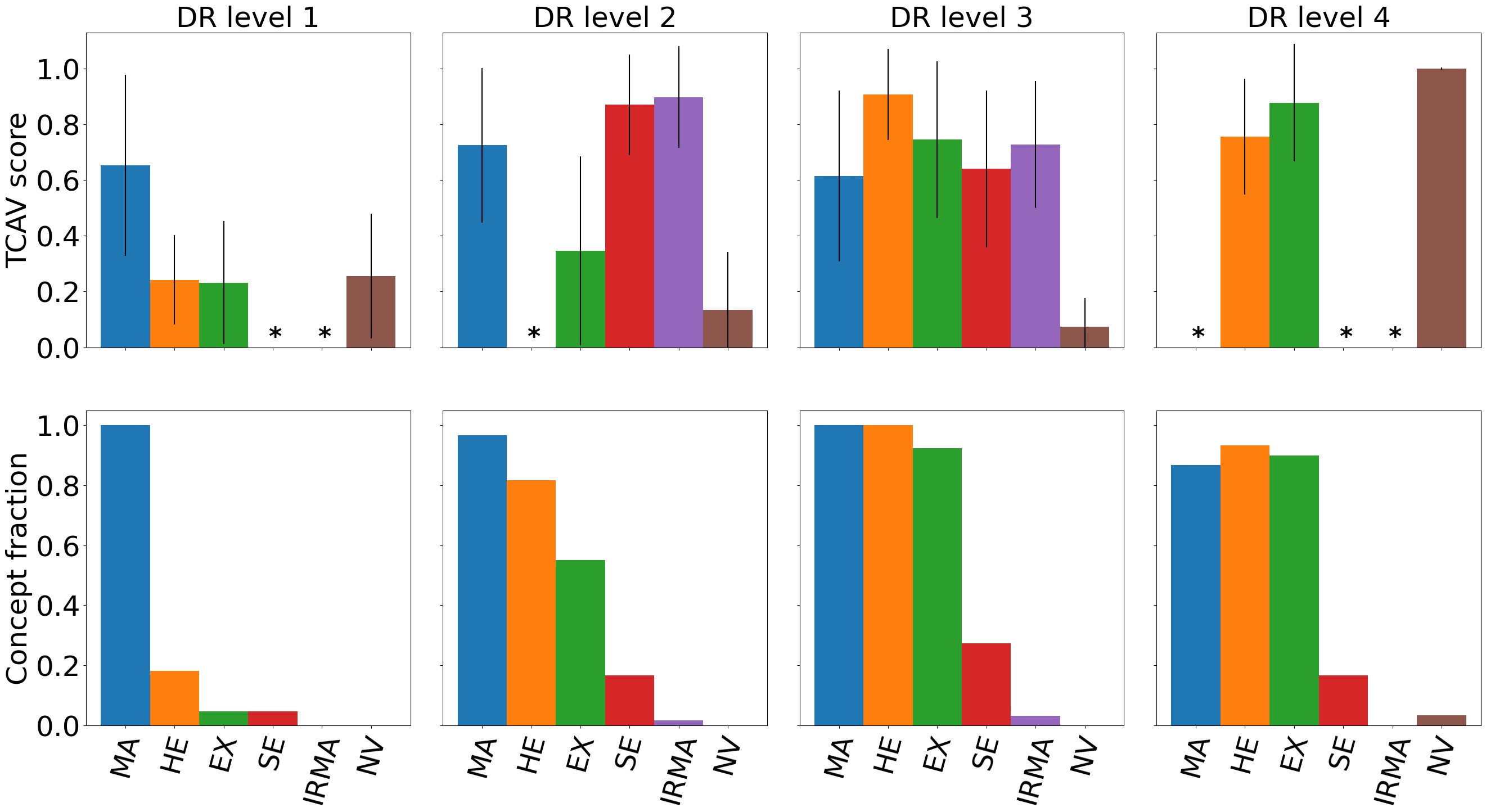}
     \caption{Upper row: \Ac{TCAV} scores for \ac{DR} levels 1 to 4, showing the mean and standard deviation for $20$ pairs of positive and negative sets for the representative test set. \textbf{$\ast$} marks insignificant concepts. Lower row: Fraction of images with concepts predicted as present in the FGADR test set by the \ac{CBM}. The values are normalized by the total number of images for each level in the test set.}
     \label{fig:CombinedTCAV_bottleneckConcepts_subplots}
\end{figure*}

Due to memory limitations, \ac{TCAV} scores were calculated on a representative subset of the combined test set consisting of $50$ randomly picked images from each \ac{DR} level. The performance for \ac{DR} prediction on the representative test set did not differ significantly from the full test set. As mentioned in~\cref{Sec:Method_TCAV}, the \ac{TCAV} concepts were either based on the full images or only the image area surrounding the medical finding(s) of interest. Masking the concept images did not generate significantly different results, thus the full concept images were used. 
\Cref{fig:CombinedTCAV_bottleneckConcepts_subplots} shows both \ac{TCAV} scores for each concept and the concept predictions with the \ac{CBM} at different \ac{DR} levels. Increasing \ac{DR} severity is associated with higher \ac{TCAV} scores for more concepts and higher concept counts. Note that the \ac{TCAV} scores and \ac{CBM} concept counts are not directly comparable. Different datasets were used, and the \ac{TCAV} scores reflect relative importance between the concepts while the \ac{CBM} counts are merely the predicted presence of concepts.

\begin{figure*}[!t]
     \centering \includegraphics[width=\textwidth]{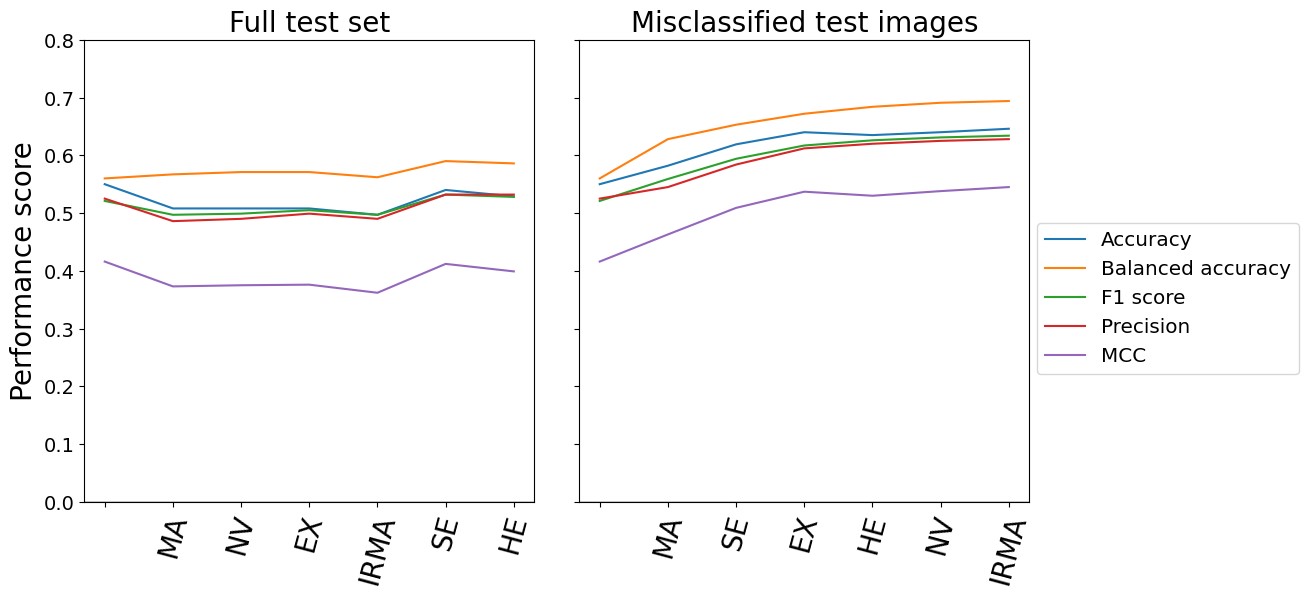}
     \caption{Performance metrics for the DR classification task during test time intervention for an increasing number of concepts. Only wrongly predicted concepts were intervened on. Left side: Results for the full FGADR test set. Right side: Results for the misclassified images in the FGADR test set.}
     \label{fig:CombinedTTI}
\end{figure*}

\begin{figure*}[!b]
     \centering
     \newcommand{\subfiguresize}{.49\columnwidth}
     \begin{subfigure}{\subfiguresize}
        \includegraphics[width=\textwidth]{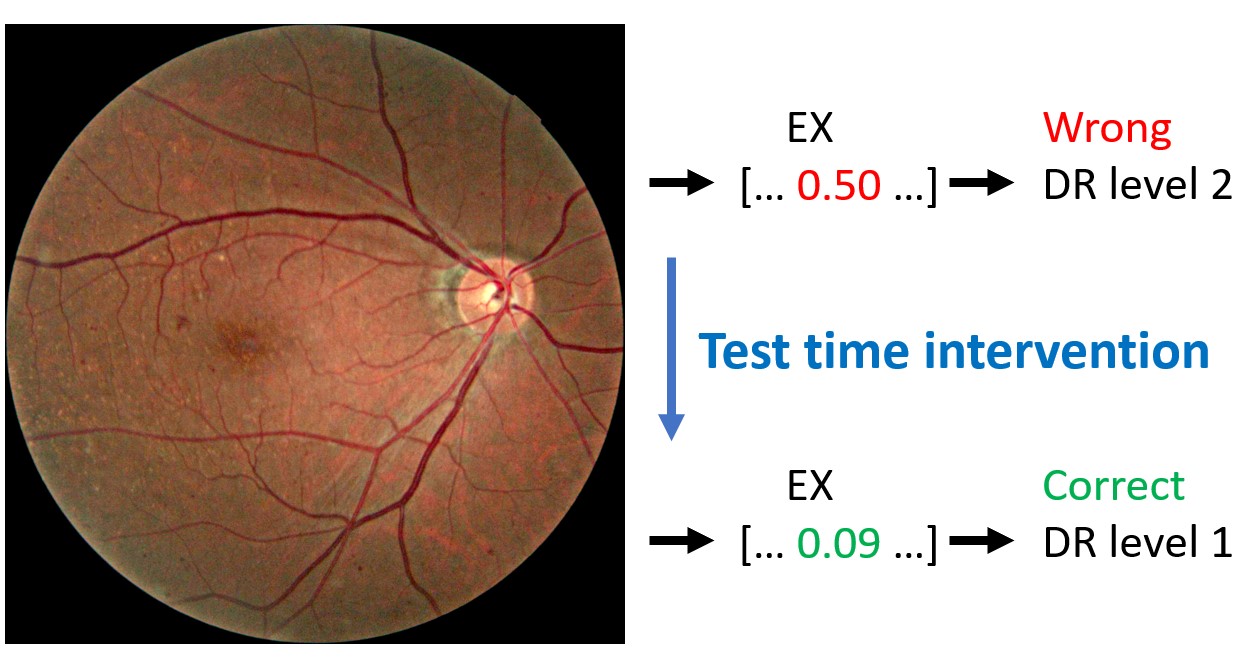}
     \end{subfigure}
     \begin{subfigure}{\subfiguresize}
        \centering
        \includegraphics[width=\textwidth]{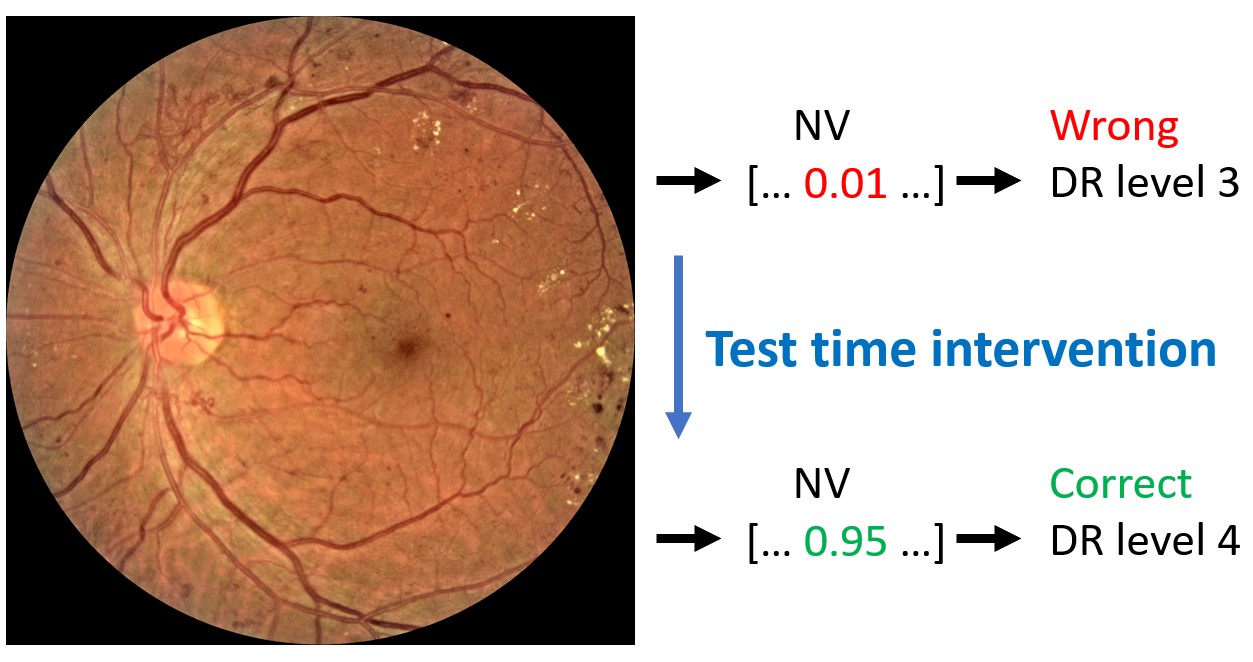}
     \end{subfigure}
        \caption{Test time intervention on selected test images with \ac{DR} levels 1 (left) and 4 (right), showing how the predicted \ac{DR} levels change. Inspired by~\cite{Koh2020BottleneckModels}.}
        \label{fig:TestTimeIntervention_selectedImages}
\end{figure*}

Test time intervention on an increasing number of concepts was performed for the \ac{CBM} trained on six concepts. The intervention concerned concepts that were wrongly predicted by the \ac{CBM}. As previously mentioned, the order of which concepts to intervene on was determined by the corresponding balanced accuracy in \ac{DR} classification following the intervention on a single concept. Consequently, the concept resulting in the best balanced accuracy on the \ac{DR} grading was included first, while the concept with the worst performance was included last. The left hand side of~\cref{fig:CombinedTTI} shows the \ac{DR} classification performance metrics on the entire \ac{FGADR} test set when concept intervention was performed incrementally. We observe that the intervention had best effect when five out of six concepts were corrected (correcting the \ac{HE} concept did not show further improvement). Even though the performance did not increase dramatically, test time intervention improved the balanced accuracy and precision compared to no intervention. Next, test time intervention was performed only on the misclassified images in the test set. As seen on the right hand side of~\cref{fig:CombinedTTI}, the performance peaked when all six concepts were corrected. The balanced accuracy increased from 56.0\% with no corrections to 69.4\%. Qualitative examples on how test time intervention affected the \ac{DR} level predictions are provided in~\cref{fig:TestTimeIntervention_selectedImages}.

\section{Discussion and Conclusion}\label{sec:Discussion}
According to the diagnostic criteria for \ac{DR} level 1 (mild \ac{DR}), \ac{MA} should be the only abnormality present~\citep{Wilkinson2003DRgrading}. This corresponds well to both \ac{TCAV} and \ac{CBM} results in~\Cref{fig:CombinedTCAV_bottleneckConcepts_subplots}, highlighting \ac{MA} as the most important concept for \ac{DR} level 1.
The \ac{IRMA}, \ac{SE}, and \ac{MA} concepts were ranked highest by \ac{TCAV} for \ac{DR} level 2, and all concepts but \ac{NV} was identified by the \ac{CBM}. This is consistent with the diagnostic criteria for \ac{MA}, \ac{HE}, and \ac{SE}. Interestingly, the \ac{IRMA} concept was ranked highest by \ac{TCAV}, even though this finding is mainly associated with \ac{DR} level 3. By inspecting the datasets more closely, several \ac{DR} level 2 images actually contained \ac{IRMA}. It is therefore reasonable that the model identifies \ac{IRMA} as important when making predictions on \ac{DR} level 2 images. Additionally, several \ac{DR} level 3 images were predicted to belong to \ac{DR} level 2. This can also partly be explained by the high \ac{TCAV} scores for the \ac{IRMA} concept.  
The \ac{TCAV} scores for \ac{DR} level 3 gave high importance to \ac{MA}, \ac{HE}, \ac{EX}, \ac{SE}, and \ac{IRMA}, and are coherent with the diagnostic criteria for this \ac{DR} level, as well as the presence of all concepts except NV for the \ac{CBM}. The presence of \ac{HE}, \ac{EX}, and \ac{SE} was increased compared to level 2, which is also expected.
Finally, for \ac{DR} level 4, \ac{TCAV} ranked \ac{NV} highest, followed by \ac{EX} and \ac{HE}, which was expected as this is the only \ac{DR} level where \ac{NV} is present. \Ac{NV} was also identified by the \ac{CBM}.
Taken together, the concept-based explanations seem to align with established medical knowledge about \ac{DR}. This is encouraging in terms of applying \ac{TCAV} and \acp{CBM} in the clinic for explaining deep neural networks for \ac{DR} grading. The next step would be to validate the usability of these concept-based XAI methods with feedback from ophthalmologists. In this current study, we compare the explanations with widely accepted guidelines for grading of \ac{DR}, but there might be other clinically relevant aspects that were not considered.

The overall ranking of the concepts were not different for the concepts based on full images and concepts based on masked images that focused on the area containing the medical findings of interest. This could mean that even though some findings are small, they are still sufficiently learnt from the full images. On the other hand, similar results could also indicate that the masking technique was not efficient enough. In order to avoid extreme deviations in the image sizes for the masked concept images, a lower limit of 520 $\times$ 520 pixels was applied. For images of low resolutions, this restriction implied that a very small part of the image was removed and that the masked and full image versions were almost the same. However, since the model was trained on images input size 620 $\times$ 620 pixels, input images with few pixels were regarded as less likely to provide useful results. Consequently, the lower pixel limit was considered the best alternative when preparing the masked concept images. The small variations between the results from full concept images and masked concept images indicate that it is sufficient to use the original images for concept generation. Further on, time is saved because we do not have to mask out the relevant findings from the images in order to get meaningful concepts.
 
\Ac{IRMA} and \ac{NV} are typical findings for severe \ac{DR}. The presence or absence of these findings is therefore expected as useful information when learning to grade \ac{DR}. As observed in~\cref{tab:resultsCombined}, the \ac{CBM} trained on four concepts (\ac{MA}, \ac{HE}, \ac{EX} and \ac{SE}) performed worse when grading \ac{DR} on the FGADR test set. Even though the model trained on four concepts received more training data and was better at predicting these four concepts than the model trained on all six concepts, missing information about \ac{IRMA} and \ac{NV} seems to negatively affect the \ac{DR} grading. This highlights the importance of high quality concept annotations for these \ac{XAI} methods.

Test time intervention showed to be a great advantage of the \ac{CBM} and resulted in more accurate model predictions. When the concepts for incorrect predictions are inspected, the user can get important information about why an image was misclassified. As an example, the \ac{DR} level $4$ image in~\cref{fig:TestTimeIntervention_selectedImages} was misclassified as \ac{DR} level $3$, because the bottleneck model missed the \ac{NV} concept. By correcting this, the image was correctly classified. 

Despite including $>33800$ images for training the combined \ac{DR} classification model, the model did not outperform earlier deep neural networks~\citep{Lakshminarayanan2021DRReview}. In this work, we combined four different datasets where the fundus images were captured using various devices and arriving from patient populations in different geographical areas. The high image diversity probably makes it challenging for a model to capture representative patterns in the data compared to more homogenous datasets. On the other hand, the diverse collection of training data could make the model robust to variations and noise in data from a real-world setting and increase its ability to generalize to new datasets. We could not identify any previous studies using the same combination of data as us, meaning that our results are not directly comparable to previously reported performance metrics. Because this study focuses on the explanation methods, the performance is regarded as sufficiently high.

Concept-based explanations are promising for increasing the understanding of \ac{DR} grading with deep neural networks. While \acp{CBM} allow for test time intervention on the concepts, these models are limited by the lack of publicly available medical datasets annotated with both concepts and target labels. For \ac{TCAV}, concepts can be defined using other data sources, meaning that the training data does not need additional concept annotations. Consequently, the model explained by \ac{TCAV} outperformed the \acp{CBM} for \ac{DR} grading.
Our results highlight a major drawback of the \acp{CBM}: Because the development dataset must be annotated with concepts and \ac{DR} level, the amount of available data is small. The requirement of training a modified \ac{CBM} also complicates direct comparison and combination with other \ac{XAI} methods. Before larger medical datasets annotated with concepts and target labels are available, \ac{TCAV} gives the best trade-off between model performance and explainability for \ac{DR} grading.


\acks{The research presented in this paper has benefited from the Experimental Infrastructure for Exploration of Exascale Computing (eX3), which is financially supported by the Research Council of Norway under contract 270053.}

%
\ethics{The work follows appropriate ethical standards in conducting research and writing the manuscript, following all applicable laws and regulations regarding treatment of animals or human subjects.}

\coi{The authors declare no conflicts of interest.}

\data{The datasets applied in the presented research are publicly available. Details regarding data and code are available on GitHub: \url{https://github.com/AndreaStoraas/ConceptExplanations\_DR\_grading}}

\bibliography{sample}



\end{document}